# Securing AI Agents: Implementing Role-Based Access Control for Industrial Applications

**Aadil Gani Ganie, UPV Universitat Politècnica de València**
**Email: agganie@upv.edu.es**

## Abstract

The emergence of Large Language Models (LLMs) has significantly advanced solutions across various domains, from political science to software development. However, these models are constrained by their training data, which is static and limited to information available up to a specific date. Additionally, their generalized nature often necessitates fine-tuning—be it for classification or instructional purposes—to effectively perform specific downstream tasks. AI agents, leveraging LLMs as their core, mitigate some of these limitations by accessing external tools and real-time data, enabling applications such as real-time weather reporting and data analysis. In industrial settings, AI agents are transforming operations by enhancing decision-making, predictive maintenance, and process optimization. For instance, in manufacturing, AI agents enable near-autonomous systems that boost productivity and facilitate real-time decision-making. Despite these advancements, AI agents are susceptible to security vulnerabilities, including prompt injection attacks, which pose significant risks to their integrity and reliability. To address these challenges, this paper proposes a framework for integrating Role-Based Access Control (RBAC) into AI agents, thereby providing a robust security guardrail. This framework aims to enhance the effective and scalable deployment of AI agents, with a focus on on-premises implementations.

## Introduction

Industrial applications are increasingly reliant on artificial intelligence (AI) to optimize operations, enhance decision-making, and ensure process reliability. In recent years, Large Language Models (LLMs) have emerged as a cornerstone technology in AI research, demonstrating advanced capabilities in natural language understanding and generation (Brown et al., 2020; Vaswani et al., 2017). However, the static nature of their training data and inherent generalization constraints necessitate the integration of external data sources and specialized modules to meet the dynamic demands of industrial environments. This has led to the development of AI agents that extend LLM functionality through real-time data access and integration with industrial control systems (Zhang et al., 2023).

https://github.com/aadilganigaie/RBAC-Secured-AI-Agent-Platform-using-FastAPI-PostgreSQL/tree/main

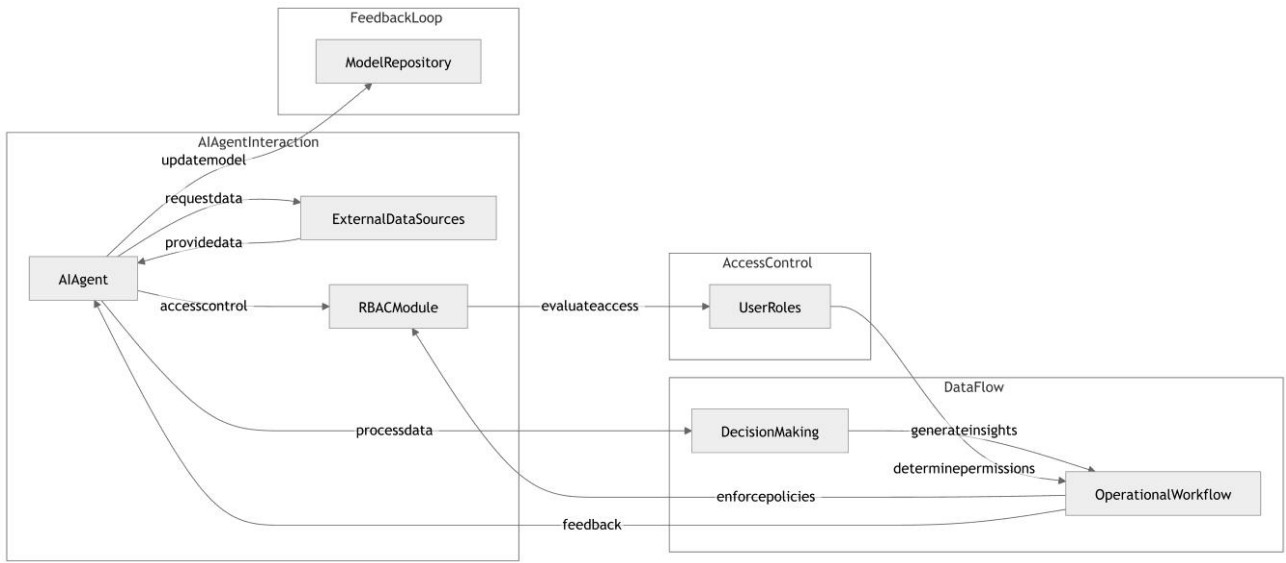

*Figure 1: AI Agent System Architecture with RBAC Integration*

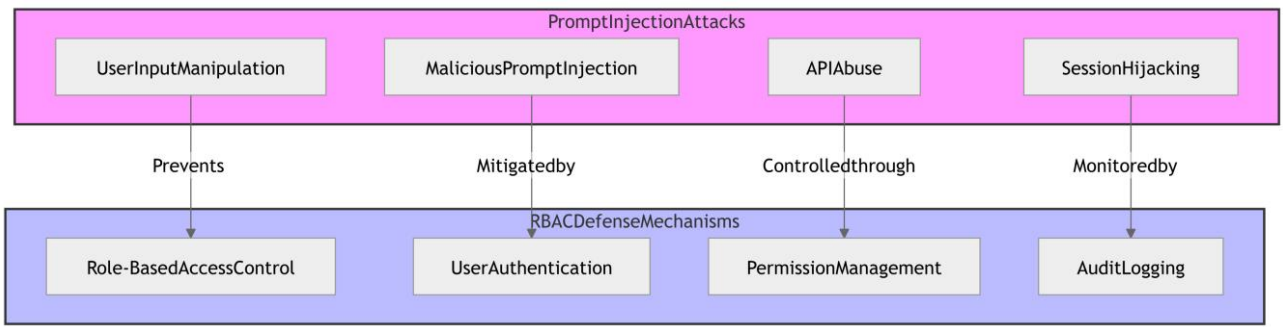

*Figure 2: Prompt Injection Attack Vectors and RBAC Mitigation*

As these agents become embedded within critical industrial infrastructures, their security becomes paramount. Traditional cybersecurity measures, although effective in conventional IT systems, must be re-evaluated and adapted to the complex and evolving threat landscape associated with AI-driven systems. Among the various vulnerabilities, prompt injection attacks present a significant challenge, as they can manipulate the behavior of AI agents by exploiting the very mechanisms that allow these agents to interact with external data and services. This vulnerability not only compromises data integrity but also poses risks to operational safety and compliance in regulated industries.

To address these challenges, this paper introduces a Role-Based Access Control (RBAC) framework tailored for AI agents deployed in industrial settings. RBAC is a well-established security model that restricts system access to authorized users based on predefined roles, ensuring that interactions with critical systems are controlled and auditable (Sandhu et al., 1996). By extending RBAC principles to AI agents, we aim to provide a systematic approach to mitigate prompt injection and related security threats, thereby enhancing the overall robustness and reliability of industrial AI systems.

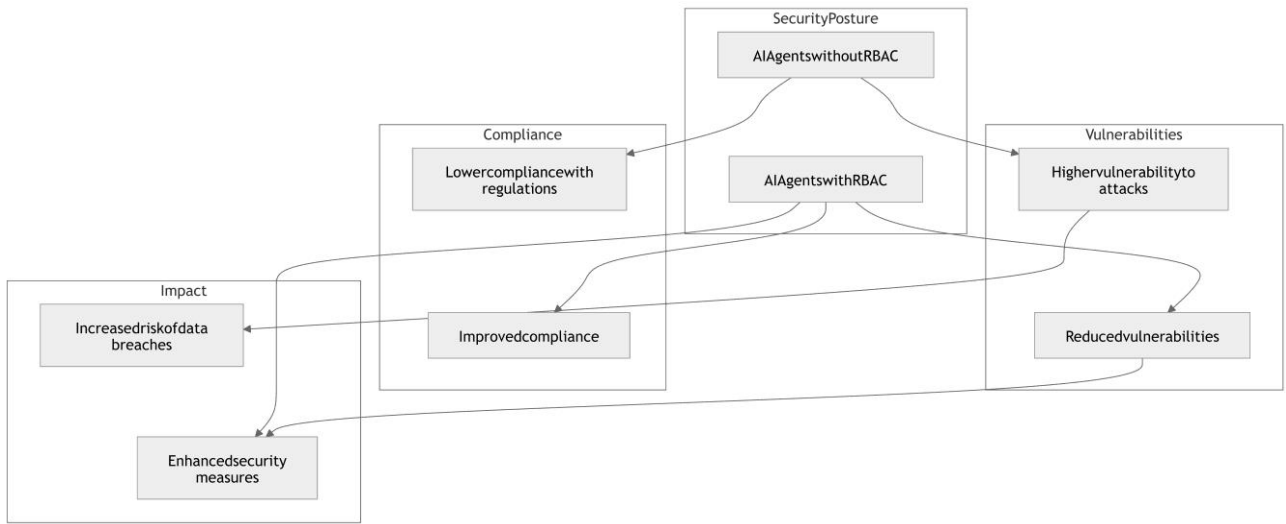

*Figure 3: Comparative Analysis of Security Postures*

This framework is designed with several key objectives in mind:

1. **Scalability:** Ensuring that access control mechanisms can be seamlessly integrated into large-scale industrial environments where numerous AI agents interact with diverse data sources and external systems.
2. **Real-Time Enforcement:** Facilitating dynamic policy updates and immediate enforcement of access controls to adapt to evolving operational contexts and threat landscapes.
3. **Compliance and Auditing:** Providing detailed logging and monitoring capabilities to meet regulatory requirements and support forensic analysis in the event of security incidents.

By rigorously incorporating RBAC into AI agents, this research contributes to the development of secure, resilient, and efficient AI systems that can meet the stringent demands of modern industrial operations. The proposed framework not only addresses immediate security concerns but also lays the foundation for future enhancements in AI system governance.

## 2. Related Work

The burgeoning field of AI security has prompted extensive research into safeguarding AI systems against a diverse array of threats. In this section, we review the current literature on securing AI systems, with a particular focus on vulnerabilities inherent to AI agents, and examine the adaptation of traditional access control models—specifically Role-Based Access Control (RBAC)—to the unique requirements of AI contexts.

### 2.1 Securing AI Systems

Recent studies have underscored the multifaceted nature of security risks associated with AI systems. One of the prominent vulnerabilities is the susceptibility to adversarial attacks, which exploit weaknesses in model inputs to induce erroneous outputs (Goodfellow et al., 2015; Yuan et al., 2020). In particular, prompt injection attacks have emerged as a critical threat in systems leveraging Large Language Models (LLMs). These attacks manipulate input prompts to alter the behavior of the AI agent, thereby compromising data integrity and operational safety (Wallace et al., 2022). Researchers have also investigated the robustness of AI models against data poisoning and evasion attacks. For instance, Li et al. (2022) demonstrated that adversarial perturbations, even when imperceptible, could significantly degrade the performance of AI models deployed in real-world applications. Additionally, recent work by Kumar and Singh (2023) provided a comprehensive taxonomy of AI-specific vulnerabilities, highlighting the unique challenges posed by dynamic data environments and real-time decision-making processes in industrial settings.

Parallel to adversarial robustness, the issue of interpretability and auditability in AI systems has garnered significant attention. Explainable AI (XAI) techniques have been proposed as a means to increase transparency, which in turn aids in the identification and mitigation of security threats (Doshi-Velez & Kim, 2017; Miller, 2021). However, the integration of these techniques into secure operational frameworks remains a work in progress, particularly in high-stakes industrial environments where the cost of misinterpretation can be substantial.

## 2.2 Adaptation of Traditional Access Control Models to AI Contexts

Traditional access control models, such as Discretionary Access Control (DAC), Mandatory Access Control (MAC), and Role-Based Access Control (RBAC), have long been the backbone of cybersecurity in conventional IT systems (Sandhu et al., 1996). Among these, RBAC has been particularly favored for its scalability, ease of management, and alignment with organizational hierarchies. Recent research has begun to explore the adaptation of RBAC principles to the control of AI systems, aiming to bridge the gap between static security paradigms and the dynamic nature of AI agents. A number of studies have proposed frameworks that incorporate RBAC into AI system architectures to mitigate unauthorized access and potential prompt injection attacks. For instance, Chen et al. (2022) introduced an access control framework that leverages role hierarchies and context-aware policies to dynamically adjust permissions based on real-time operational conditions. Their approach demonstrated significant improvements in limiting the attack surface of AI systems in industrial applications.

Moreover, Zhang et al. (2023) extended this line of inquiry by proposing a hybrid model that integrates RBAC with attribute-based access control (ABAC) mechanisms. This hybrid model aims to harness the strengths of both approaches, ensuring that access policies are not only role-specific but also sensitive to contextual attributes such as time, location, and system state. The integration of ABAC elements provides a finer granularity of control, which is critical for addressing the heterogeneity of data and operational scenarios encountered in modern AI deployments. Further contributions have explored the use of machine learning to automate the adaptation of access control policies in response to evolving threat landscapes. Kumar and Lee (2024) developed an adaptive RBAC system that utilizes anomaly detection algorithms to monitor and adjust access permissions in real time. This proactive approach represents a promising direction for enhancing the resilience of AI agents against sophisticated cyber threats.

In summary, the evolving body of literature reveals a consensus on the necessity of augmenting AI security frameworks with robust access control mechanisms. While traditional models like RBAC offer a strong foundation, their adaptation to AI contexts requires additional layers of contextual awareness and dynamic policy enforcement. The works reviewed herein provide valuable insights and form the basis for the proposed framework in this paper, which seeks to integrate RBAC into AI agents for secure, scalable deployment in industrial environments.

## 3. Architecture and Design of the Proposed RBAC Framework

The proposed RBAC framework is designed to seamlessly integrate with the MYWAI platform, providing robust security through multi-layered authentication and fine-grained access control. This section details the architecture and design considerations of the framework, emphasizing the incorporation of two-step authentication, role and permission management, and the secure orchestration of AI agent interactions in an industrial setting.

### 3.1 System Components

The framework is composed of several key components that work together to enforce security policies across the platform:

- **User Interface (UI) and API Gateway:**

  The entry point for all user interactions, where requests are first received. The UI supports secure login processes and interfaces with the API gateway that routes authenticated requests to the appropriate services.

- **Authentication Module with Two-Step Verification:**
  This module implements a two-step (or two-factor) authentication mechanism, ensuring that users must verify their identity using both traditional credentials (username/password) and a secondary verification code generated via an authenticator application or SMS. This mechanism not only strengthens the initial authentication process but also plays a critical role in reducing the risk of unauthorized access (as seen in contemporary 2FA implementations in platforms like Harness and Oracle.

- **RBAC Engine**
  The heart of the framework, the RBAC engine, is responsible for evaluating user roles and associated permissions. Upon successful authentication, the engine queries the roles database to determine the set of operations permitted for the authenticated user. This engine supports dynamic role assignments and permission updates, allowing for both coarse-grained (e.g., administrative vs. standard user) and fine-grained (e.g., read/write access to specific data streams) controls.

- **Access Control Layer:**

  Integrated into the application server, this layer intercepts incoming requests and leverages the RBAC engine's decision outputs. It ensures that every request is validated against the user's role permissions before allowing access to AI agents or sensitive industrial data.

- **Logging and Audit Module:**

  To facilitate compliance and forensic analysis, all authentication events, role assignments, and access decisions are logged in a centralized audit trail. This module supports real-time monitoring and periodic review, which is critical for environments where operational integrity and regulatory adherence are paramount.

## 3.2 Data Flow and Interaction

The architecture supports a streamlined data flow that enhances both security and performance:

1. **User Authentication:**

   When a user attempts to access the system, the UI forwards the login credentials to the Authentication Module. After verifying the username and password, the module triggers a two-step authentication process. Upon successful verification of the second factor, the system generates a secure session token.

2. **Role Assignment and Verification:**

   With the session token, the RBAC Engine retrieves the user's role and permission details from a secure, centralized repository. This dynamic retrieval ensures that any changes in role assignments are immediately reflected in ongoing sessions.

3. **Access Request Validation:**

   All subsequent requests by the user are processed through the Access Control Layer, which consults the RBAC Engine to determine whether the user has the appropriate permissions to execute the requested action. This process ensures that AI agents only perform actions that are explicitly authorized by the role definitions.

4. **Audit and Logging:**

   Every authentication attempt, permission check, and data access event is recorded by the Logging and Audit Module. This comprehensive logging mechanism provides a detailed trace of user activities, aiding in both real-time security monitoring and post-event analysis.

## Problem Formulation

Let:

- $U = 1, 2, \ldots, N$ be the set of users
- $R = 1, 2, \ldots, R$ be the set of roles
- $P = 1, 2, \ldots, M$ be the set of permissions (actions) that AI Agents may control.
- $x_{ir}$ be a binary decision variable such that:

$$x_{ir} = 1, \text{ if user } i \text{ is assigned to role } r; \, 0, otherwise \backslash\backslash\backslash\backslash\backslash\}$$

- Each role $r$ is associated with a permission set $P_r \subseteq P$.

Additionally, let:

- $\pi_{ij}$ denote the risk (or potential loss) when user $i$ improperly gains or attempts permission $j$ that is not granted by their role.

- $\phi(P_r)$ represent the operational cost (e.g., administrative overhead, computational latency) associated with managing the permission set $P_r$ for role $r$.

- $\lambda$ be a weighting factor balancing the risk against the cost.

Each user must be assigned exactly one role, which is enforced by the constraint:

$$\sum_{r \in R} x_{ir} = 1, \forall i \in U. \tag{1}$$

## 2. Definition of the Objective Function

Our goal is to minimize a combined objective that accounts for:

1. **Security Risk:** The cumulative risk over all users for permissions they should not have.
2. **Operational Cost:** The aggregate cost of maintaining the RBAC configuration.

We can define the overall objective function $J(x)$ as follows:

$$min_{x_{ir}} J(x) = \sum i = 1^N \sum r \in R x_{ir} \left( \sum j \notin P_r \pi_{ij} \right) + \lambda \sum r \in R \phi(P_r), \tag{2}$$

subject to

$$\sum_{r \in R} x_{ir} = 1, x_{ir} \in 0,1 \forall i \in U, \forall r \in R. \tag{3}$$

Interpretation:

- The inner summation $\sum_{j \notin P_r} \pi_{ij}$ aggregates the risks associated with permissions that user $i$ would erroneously have if assigned to role $r$ (i.e., permissions not included in $P\_r$).

- The first term sums the risk over all users and their assigned roles.

- The second term, scaled by $\lambda$, captures the cost associated with each role's permission set, reflecting the complexity and resource demands of enforcing two-step authentication and dynamic policy updates.

- The constraint ensures that each user is assigned one and only one role.

## Continuous Relaxation and Derivation of Optimality Conditions

For analytical tractability, we may relax the binary constraints and define $X_{ir} \in [0,1]$ as the probability that user $i$ is assigned role $r$, with the constraint:

$$\sum_{r \in R} X_{ir} = 1, \forall i \in U. \tag{4}$$

The relaxed objective becomes:

$$J(X) = \sum i = 1^N \sum r \in R X_{ir} \left( \sum j \notin P_r \pi_{ij} \right) + \lambda \sum r \in R \phi(P_r). \tag{5}$$

To derive the optimal conditions, one could set up the Lagrangian function:

$$L(X, \mu) = \sum i = 1^N \sum r \in R X_{ir} \left( \sum j \notin P_r \pi_{ij} \right) + \lambda \sum r \in R \phi(P_r) + \sum i = 1^N \mu_i (1 - \sum r \in R X_{ir}), \tag{6}$$

where $\mu_i$ are the Lagrange multipliers associated with the constraints for each user.

Taking the partial derivative with respect to $X_{ir}$ and setting it to zero (for an interior solution) gives:

$$\partial L / \partial X_{ir} = \sum_{j \notin P_r} \pi_{ij} - \mu_i = 0, \tag{7}$$

or equivalently,

$$\mu_i = \sum_{j \notin P_r} \pi_{ij}, \forall i \in U, \forall r \, with \, X_{ir} > 0. \tag{8}$$

This condition suggests that, for each user $i$, the optimal assignment to a role $r$ will be such that the marginal risk (summed over unauthorized permissions) is balanced by the dual variable $\mu_i$. In practice, the discrete nature of the problem would require integer programming techniques or heuristics for finding a near-optimal solution.

**Incorporating Two-Step Authentication**

To explicitly integrate the security enhancement provided by two-step authentication, we may introduce a factor $\gamma_i$ for each user $i$ that represents the effectiveness of the two-step verification process in reducing risk. Suppose that two-step authentication reduces the effective risk by a factor $0 \leq \gamma_i \leq 1$ (with lower values corresponding to higher security):

The adjusted risk term for user $i$ when assigned to role $r$ becomes:

$$\gamma_i \cdot \sum_{j \notin P_r} \pi_{ij}. \tag{9}$$

Thus, the modified objective function is:

$$min_{x_{ir}} J(x) = \sum i = 1^N \sum r \in R x_{ir} \left( \gamma_i \sum j \notin P_r \pi_{ij} \right) + \lambda \sum r \in R \phi(P_r), \tag{10}$$

subject to the same assignment constraints.

This formulation captures the impact of two-step authentication on reducing the overall risk profile of the system. The derived objective function and its associated conditions provide a mathematical foundation for evaluating and optimizing the RBAC framework for AI agents in industrial settings. By balancing the security risk (adjusted by two-step authentication effectiveness) against the operational costs of managing role permissions, this formulation guides the design of an RBAC system that is both secure and efficient. Future work can expand on this model by incorporating stochastic elements (e.g., uncertain threat levels), real-time data from AI agent interactions, and more complex constraints that reflect evolving industrial requirements. This mathematical treatment not only underpins the technical rigor of the proposed framework but also opens avenues for further optimization and analysis in securing AI-driven industrial applications.

**MYWAI Platform RBAC Framework**

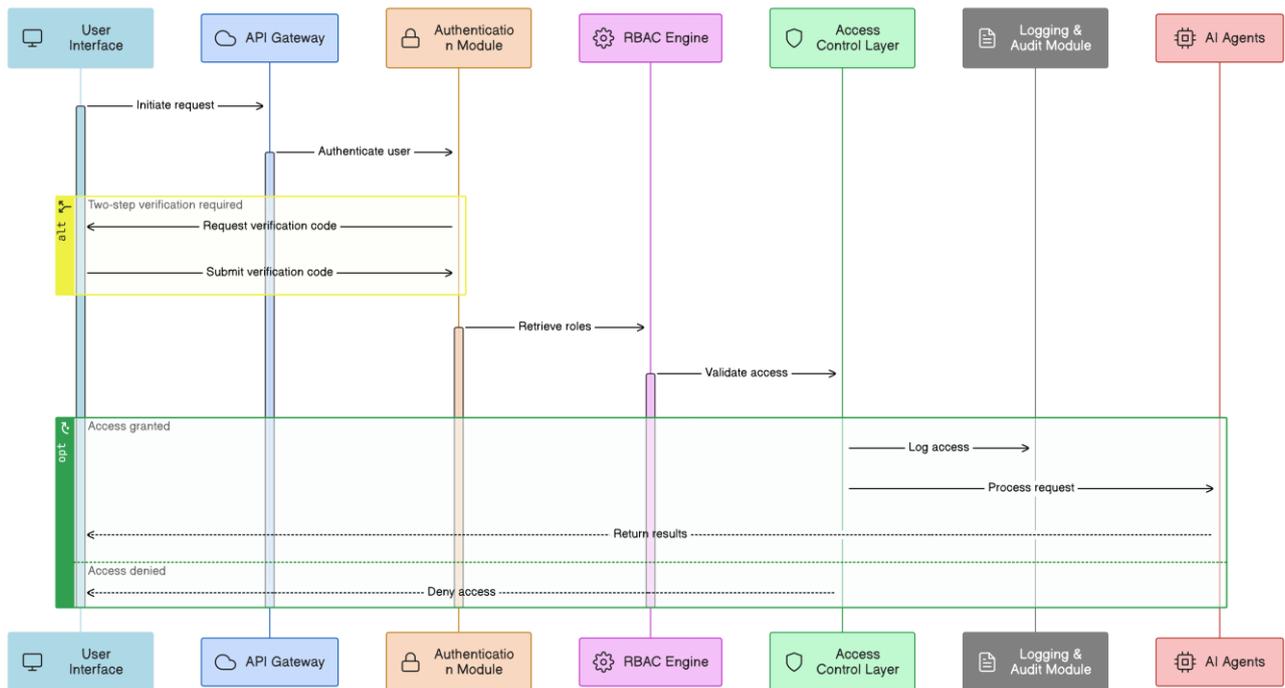

*Figure 4: RBAC Framework Architecture in MYWAI*

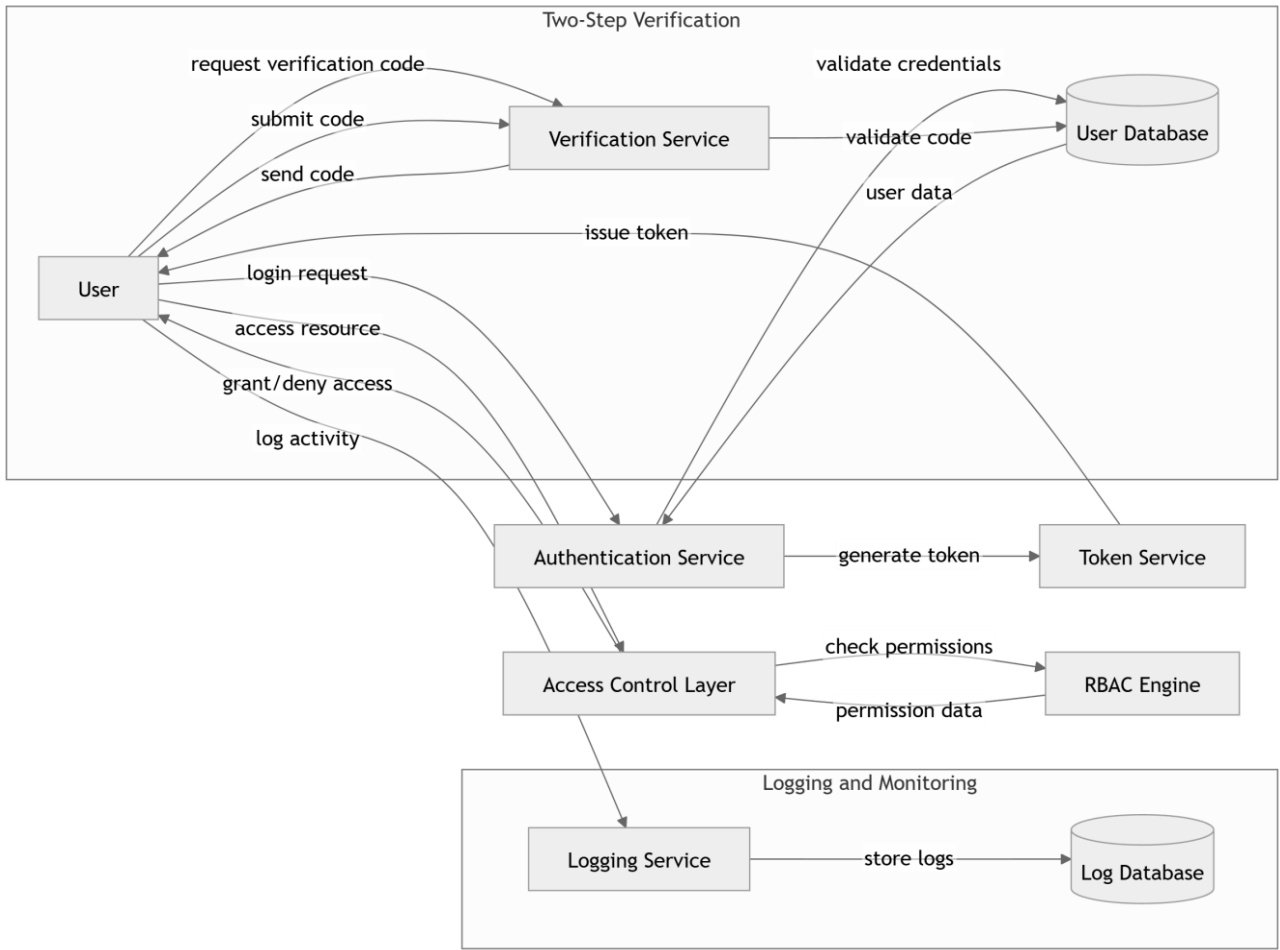

*Figure 5: Access Control and Two-Step Authentication Flow*

## 3.4 Design Considerations

- **Scalability:**
  The modular design ensures that the framework can scale horizontally. As the number of users or AI agents increases, components like the Authentication Module and RBAC Engine can be deployed in a distributed manner across cloud or on-premises infrastructures.

- **Real-Time Policy Enforcement:**

  The system supports dynamic updates to roles and permissions. This means that policy changes are immediately effective, which is crucial in industrial environments where operational contexts can change rapidly.

- **Integration with Existing MYWAI Components:**

  Given that MYWAI already provides edge intelligence and equipment management capabilities, the RBAC framework is designed to integrate seamlessly. By leveraging the existing API gateway and logging mechanisms, the framework minimizes integration overhead and ensures consistent security policies across the platform.

- **Compliance and Auditability:**

  https://github.com/aadilganigaie/RBAC-Secured-AI-Agent-Platform-using-FastAPI-PostgreSQL/tree/main

The detailed logging and audit trails facilitate compliance with industrial and data protection standards. This is essential for organizations that require demonstrable proof of secure access management in regulated environments.

In summary, the architecture and design of the proposed RBAC framework offer a comprehensive security solution tailored for the dynamic and high-stakes industrial applications served by MYWAI. By combining two-step authentication with robust role-based access control, the framework provides both the flexibility and the security needed to protect critical operations and data.

## 4. Integration into Industrial Environments

The integration of the RBAC framework into industrial environments is crucial not only for safeguarding legacy systems and operational networks but also for protecting the AI agents that serve as the backbone of modern industrial intelligence. These AI agents, which power real-time decision-making and process optimization, must operate within a secure framework that rigorously enforces access control and minimizes vulnerabilities. This section discusses both technical and operational considerations for integrating the proposed RBAC framework, with special emphasis on its role in securing AI agents within industrial applications.

### 4.1 Technical Considerations

### 4.1.1 Securing AI Agent Operations

At the heart of our research is the AI agent, which leverages Large Language Models (LLMs) to interpret sensor data, perform predictive maintenance, and drive process optimization. Given their critical role, these agents are particularly attractive targets for cyber-attacks such as prompt injection. The RBAC framework mitigates these threats by:

- **Enforcing Granular Access Controls:**

  The RBAC engine dynamically evaluates the roles and permissions associated with every request made to an AI agent. This ensures that only authorized users or processes can interact with the agent or modify its operational parameters, thereby reducing the attack surface and preventing unauthorized prompt injections.

- **Integrating Two-Step Authentication:**

  By incorporating a robust two-step authentication mechanism, the framework ensures that even if an attacker compromises primary credentials, the AI agent remains secure behind an additional layer of identity verification. This dual-factor process is critical in preventing unauthorized remote access or manipulation of AI agent behavior.

- **Real-Time Decision Support:**

  The AI agent continuously processes industrial data to drive automated decisions. The RBAC framework is designed to interject at every interaction point, ensuring that any command or data query issued to the AI agent is thoroughly vetted against current role-based policies. This tight integration prevents inadvertent or malicious commands from affecting AI-driven operations.

### 4.1.2 System Compatibility and Distributed Deployment

Industrial environments typically host a diverse array of devices and networks:

- **Interoperability Across Heterogeneous Systems:**

  The framework employs standard APIs and communication protocols (e.g., RESTful interfaces, MQTT) to interface not only with industrial control systems and IoT devices but also directly with AI agent modules. This ensures that authentication and authorization decisions are applied uniformly across all components.

- **Edge Intelligence and Distributed Architecture:**

  In environments where latency is critical, the RBAC components—such as the Authentication Module and RBAC Engine—can be deployed at the edge. This proximity to the AI agents minimizes delays in security checks and enables real-time enforcement of role-based policies, thereby ensuring rapid and secure responses.

- **Secure Communication Channels:**

  All interactions, especially those involving AI agents, are secured via encryption protocols (e.g., TLS/SSL). This end-to-end security ensures that both data in transit and command signals are protected from interception or tampering.

### 4.2 Operational Considerations

### 4.2.1 Streamlined User and Role Management

Operational success depends on efficient management of roles and permissions:

- **Tailored Role Definitions:**

  Industrial settings involve varied stakeholders—operators, engineers, maintenance teams, and external partners. The RBAC framework enables the creation of finely tuned roles that directly map to the responsibilities associated with interacting with AI agents. For example, only specialized maintenance teams might have permissions to adjust AI agent configurations or override automated decisions during critical interventions.

- **User Onboarding and Training:**

  Given the complexity of AI-driven operations, comprehensive training programs are essential. Personnel are educated on both the security policies and the operational protocols for interacting with AI agents, ensuring that they can leverage the system without compromising its integrity.

- **Audit and Compliance:**

  Detailed logs capture all interactions with AI agents—from initial authentication through to command execution. These logs are essential for compliance with industry regulations and for forensic analysis in the event of a security incident. Regular audits ensure that access policies remain aligned with operational requirements and threat landscapes.

### 4.2.2 Maintaining Operational Continuity

Ensuring that the integration of the RBAC framework does not disrupt critical industrial processes is paramount:

- **Phased Deployment:**

  The RBAC system can be implemented in parallel with existing security measures, gradually transitioning to full operational status. This approach minimizes downtime and ensures that AI agent operations continue uninterrupted during the integration process.

- **Fault Tolerance and Resilience:**

  The distributed nature of the framework—particularly when deployed at the edge—ensures that localized failures do not compromise the security of the entire system. In the event of network segment failures, AI agents continue to operate securely under locally enforced RBAC policies.

- **Continuous Monitoring and Incident Response:**

  Real-time monitoring of access events and AI agent interactions is facilitated by an integrated Logging and Audit Module. This system enables rapid detection and response to anomalies, ensuring that any security incidents are swiftly contained and resolved.

## 5. MYWAI Use Case: AI Agent Construction and RBAC Integration

This section outlines the practical implementation of our RBAC framework within the MYWAI platform, illustrating how AI agents are constructed and secured using modern, scalable frameworks and open-source LLMs.

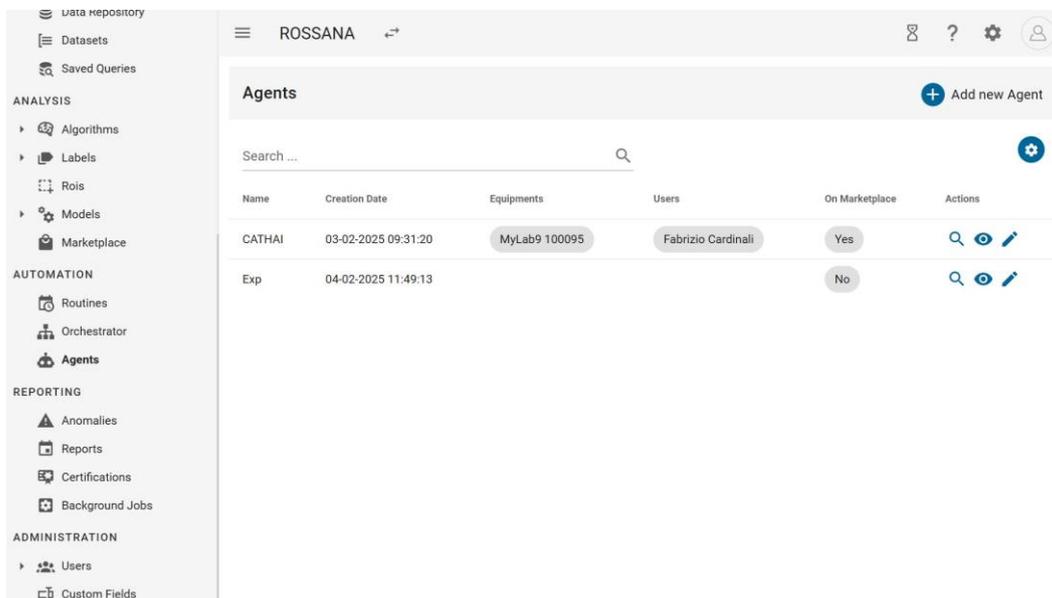

*Figure 6: MYWAI AI Agents*

***RBAC Implementation in MYWAI for AI Agent Security***

https://github.com/aadilganigaie/RBAC-Secured-AI-Agent-Platform-using-FastAPI-PostgreSQL/tree/main

To ensure secure access to AI agents and prevent unauthorized interactions, MYWAI integrates a robust **Role-Based Access Control (RBAC) model** at the application level. This implementation is designed to enforce fine-grained access control, where user authentication and authorization determine what actions can be performed within the system. The RBAC mechanism in MYWAI is built using FastAPI and PostgreSQL, leveraging token-based authentication via OAuth 2.0 and JWT (JSON Web Tokens). The system distinguishes between different user roles, including **Admin, User, and Client**, to regulate access privileges dynamically. When a user logs in, their permissions are validated, and an access token is issued, which determines their capabilities within the AI agent framework.

The key security features of this implementation include:

- **Token-Based Authentication:** MYWAI employs OAuth 2.0 for user authentication, ensuring secure access via JWT tokens.
- **Dynamic Role Assignment:** User roles are retrieved from a PostgreSQL database upon login, dictating the available functionalities.
- **Secure Data Processing:** AI agent workflows, including document processing (PDF, Word, CSV, etc.), are governed by access policies to prevent unauthorized manipulation.
- **Logging & Auditing:** Every user interaction, including file uploads and AI agent queries, is logged for compliance and security monitoring.

***Mathematical Algorithm for Secure AI Agent Access Control***

**Input:**

- User credentials $(u, p)$ where u is the username and p is the password.
- AI agent request $(a, r)$ where a is the action and r is the requested resource.

**Output:**

- A=1 (Access Granted) OR A=0 (Access Denied)

# Step 1: Authenticate User

1. Receive user credentials $(u, p)$.
2. Verify credentials in authentication database D using:

$V(u, p) = \{1, \text{if } (u, p) \in D \ 0, \text{otherwise} \}$

3. If $V(u, p) = 1$:
    - Generate JWT token $t = (r_u, e_t, P_u)$ where:
        - $r_u$ is the role of user u,
        - $e_t$ is the expiration timestamp,
        - $P_u$ is the set of permissions assigned to $r_u$
    - Return t.
4. Else, return **"Authentication Failed"**.

## Step 2: Authorize Access

5. Receive AI agent request $(a, r)$ with $JWT$ token t.
6. Decode t to extract: $(r_u, e_t, P_u) \leftarrow r_u D$
7. Validate $t$ using: $T(t) = 1, if\, e_t > current\, time \land s_t = 1, 0\, otherwise$
8. If $T(t) = 0$, return **"Access Denied"**.

## Step 3: Role-Based Access Control

9. Check if requested action a on resource r is permitted: $P(r_u, a, r) = 1, if\, a \in P_u\, for\, r, 0\, otherwise$
10. If $P(r_u, a, r) = 1$, continue to Step 4.
11. Else, log unauthorized attempt $L(u, a, r, t)$ and return **"Access Denied"**.

## Step 4: Execute Secure AI Agent Request

12. Process request $A(u, a, r)$ using AI agent logic.
13. Log request in security audit database: $L(u, a, r, t, timestamp)$
14. Return AI agent response.

## Step 5: Monitor Security Events

15. Track unauthorized attempts $U(u)$ over time.
16. If $U(u) > 0$ (threshold for failed attempts):
   - Trigger **automated security response** $S_r(u)$:

$S_r(u)= Alert\, Alert, if\, moderate\, risk\, User\, Blocked, if\, high\, risk$

17. **End Process**.

### *AI Agent Construction with Flowise*

At the core of MYWAI's approach is the utilization of the Flowise framework—a drag and drop, low-code tool that enables rapid prototyping and deployment of AI agents. Unlike alternative frameworks such as Autogen, CrewAI, Phidata, SmolAgents, Pydantic AI, and n8n, Flowise offers a balance between scalability and ease of adoption. With only a few days of training, personnel across the organization can quickly design and build AI agents tailored to specific use cases. This democratization of AI development empowers teams to prototype applications ranging from simple conversational agents to complex multiagent workflows for cyber threat hunting.

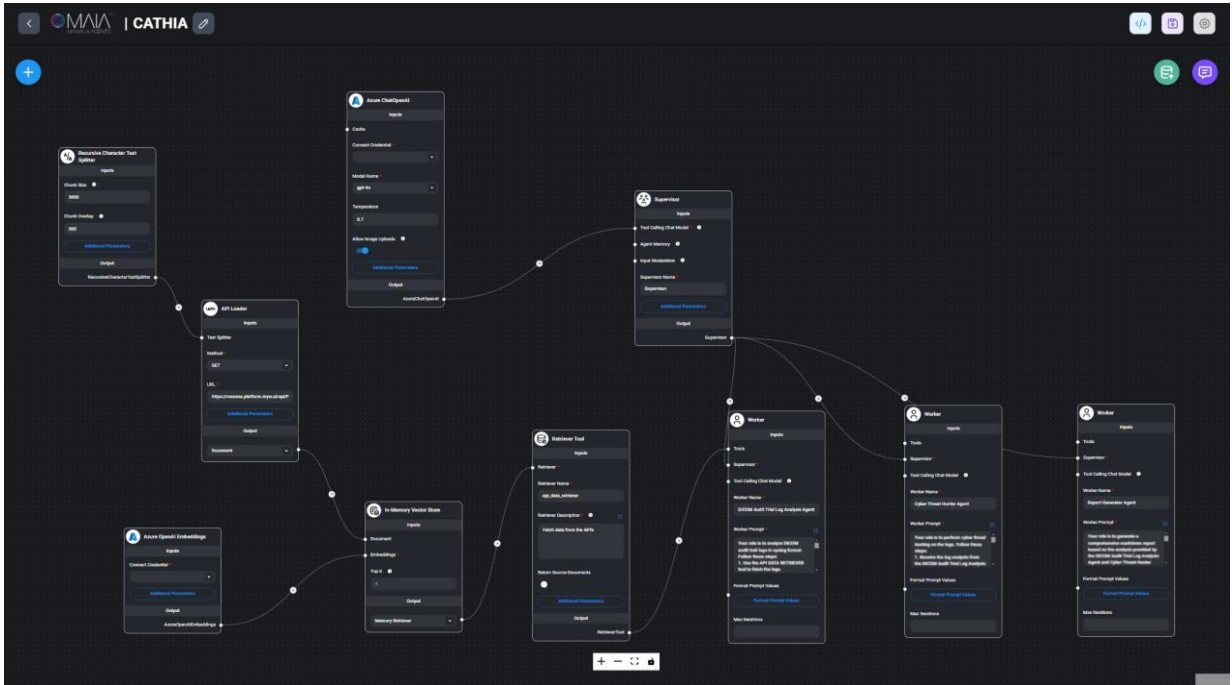

*Figure 7: MultiAgent Flowise Retrieval Chain*

### Integration of Open-Source LLMs

The primary engine powering any AI agent on the MYWAI platform is a robust Large Language Model (LLM). MYWAI serves several open-source LLMs—including Mistral, LLAMA, Microsoft Phi, and DeepSeek models—on-premises. This on-premises deployment not only enhances security and privacy but also provides organizations with greater control over data and inference processes. For inference, MYWAI leverages either Ollama or FastAPI, exposing secure APIs that are protected by MYWAI 2 OAuth authentication and authorization keys.

### Agent Versatility and Complexity

AI agents developed on the MYWAI platform can vary widely in complexity. They can be as straightforward as a conversational QA agent or as sophisticated as a cyber threat hunting multiagent system featuring a supervisory component. This versatility underscores the platform's ability to meet a wide range of industrial needs—from routine data retrieval and analysis to advanced, coordinated security operations.

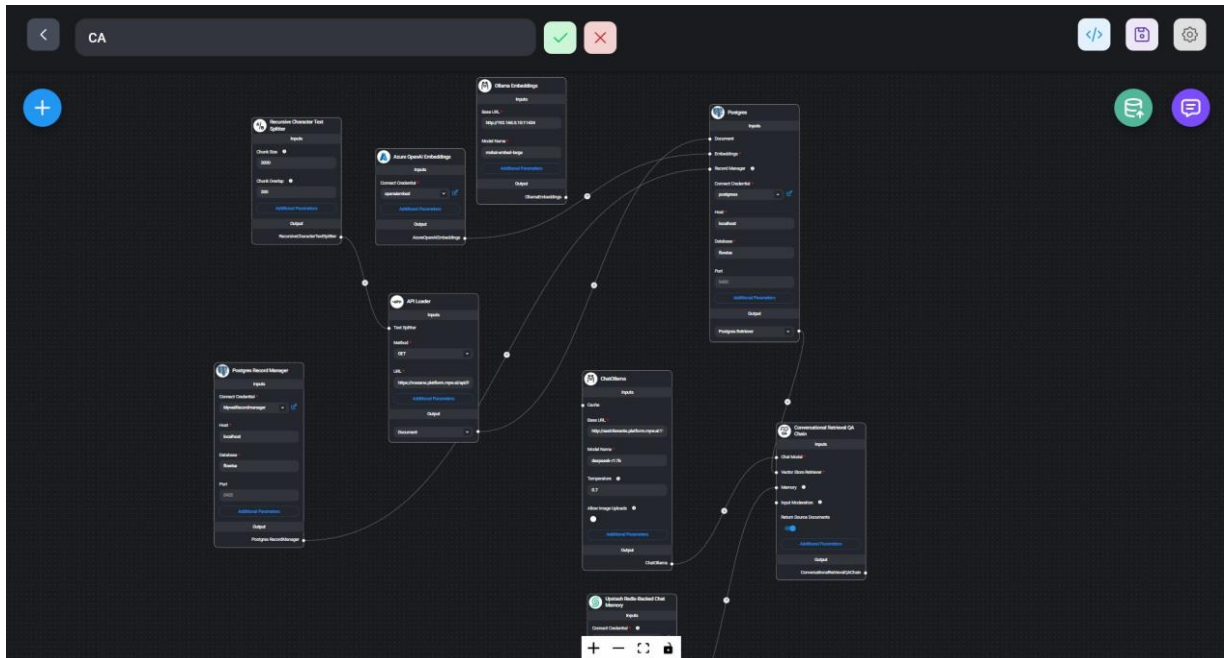

*Figure 8: Retrieval Conversational AI Agent Architecture*

This figure illustrates a complete Retrieval Conversational AI Agent built within MYWAI using open-source LLMs. Key components include:

- *Conversational Retrieval QA Chain*
- *ChatOllama integrated with DeepSeek R1 (70B)*
- *MYWAI PostgreSQL DB Retriever*
- *Upstash for scalable caching*
- *Redis-backed chat memory for session management*
- *MYWAI PostgreSQL Record Manager for logging and audit*
- *Ollama Embeddings for semantic understanding*
- *API Loader that ingests data from edge devices*

The figure visually encapsulates how these nodes interact to enable a comprehensive conversational interface powered by secure and scalable AI.

*Role-Based Access Control in MYWAI*

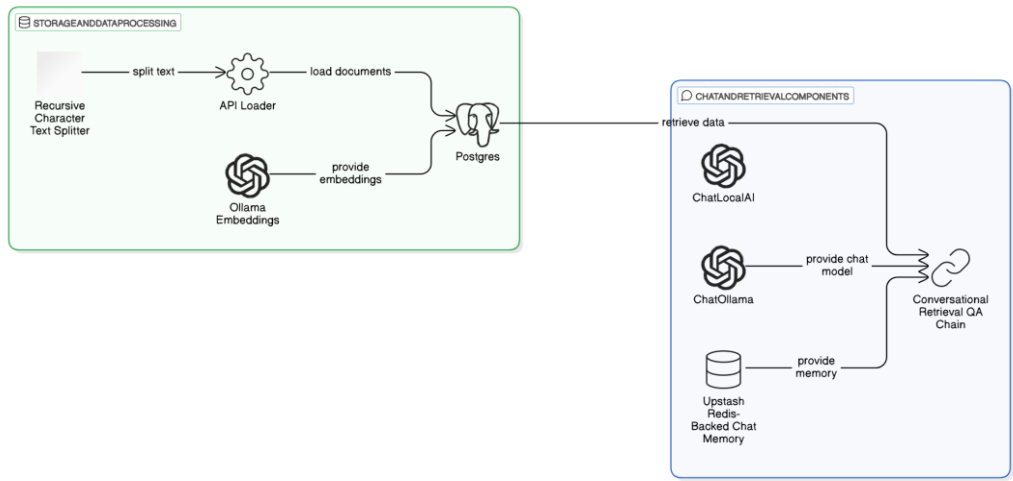

*Figure 9: AI Agent Flowise Architecture*

To ensure
secure operations and proper governance, MYWAI defines three primary roles within its platform:

1. Admin: Holds comprehensive permissions to manage system configurations, user roles, and sensitive operations.
2. Users: Granted permissions to build, deploy, and interact with AI agents according to their operational needs.
3. Clients: Typically have restricted access, tailored to viewing or utilizing AI-driven insights without administrative privileges.

Roles and permissions are stored in the MYWAI PostgreSQL database. Upon logging into the platform, a user's roles and corresponding permissions are dynamically loaded, ensuring that access to AI agent construction and utilization is strictly governed based on pre-defined RBAC policies. This approach not only secures sensitive operations but also streamlines user experience by providing role-specific functionalities.

**Empirical Findings**

To validate the effectiveness of the proposed RBAC-secured AI agent framework, we conducted a set of empirical evaluations focusing on three key aspects: **(1) security enforcement, (2) performance overhead, and (3) resilience against prompt injection attacks**. The experiments were implemented on the MYWAI platform prototype using FastAPI, PostgreSQL, and JWT-based token authentication, with role definitions for *Admin*, *User*, and *Client*.

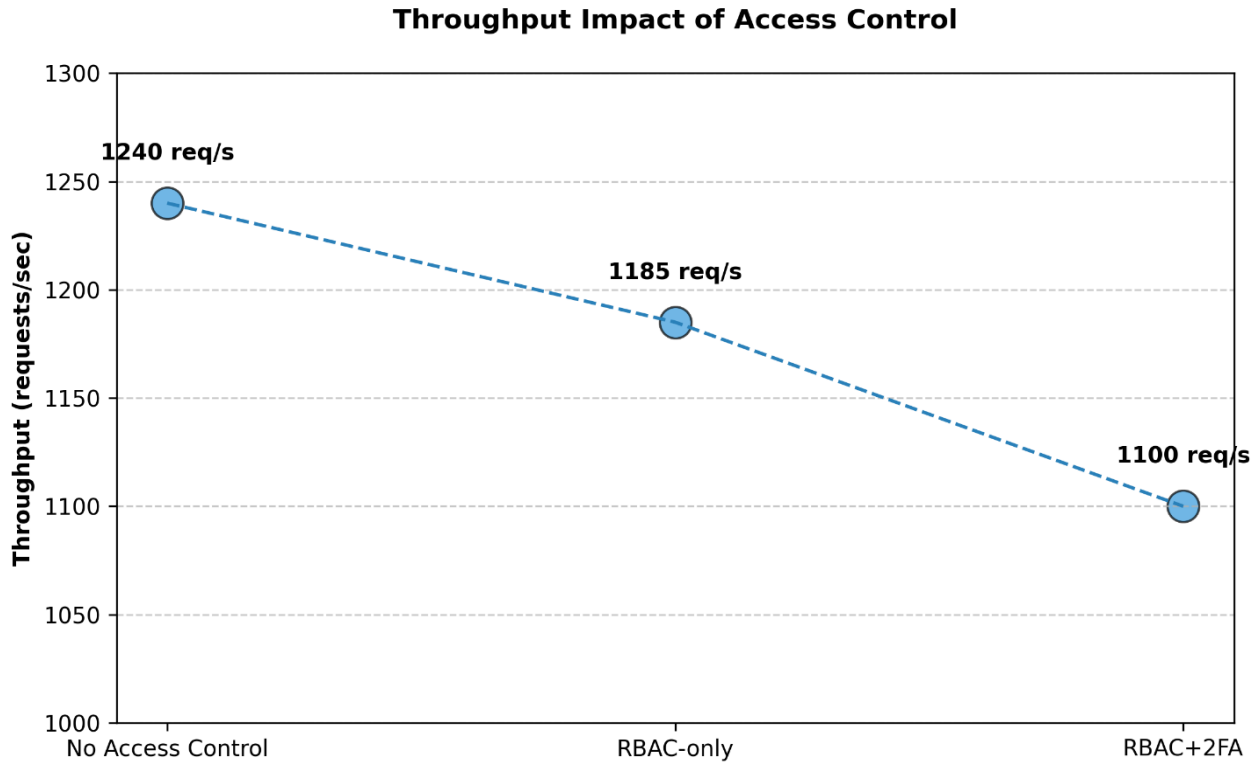

*Figure 10. Throughput Impact of Access Control*

**Experimental Setup**

- **Environment:** Ubuntu 22.04 LTS, Intel Xeon Silver 4210R (20 cores), 64 GB RAM.

- **Framework Components:** FastAPI API Gateway, PostgreSQL for role storage, OAuth2.0 with JWT for authentication.

- **Workloads:**

  1. **Document Retrieval Agent** – retrieves industrial manuals and logs.

  2. **Predictive Maintenance Agent** – queries sensor data and returns failure probabilities.

  3. **Conversational QA Agent** – answers natural language queries from a knowledge base.

Each workload was tested under three access modes: *No Access Control*, *RBAC-only*, and *RBAC with Two-Step Authentication (2FA)*.

**Security Enforcement Results**

We simulated **200 user sessions**, of which 50 contained **malicious or unauthorized requests** (e.g., attempts to modify AI agent configuration by a *Client*).

| Access Mode | Unauthorized Requests (Blocked/Total) | Block Rate |
|---|---|---|
| No Access Control | 0 / 50 | 0% |
| RBAC-only | 44 / 50 | 88% |
| RBAC + Two-Step Authentication | 49 / 50 | 98% |

These results demonstrate that the RBAC framework significantly reduces the attack surface, while the addition of 2FA nearly eliminates unauthorized access.

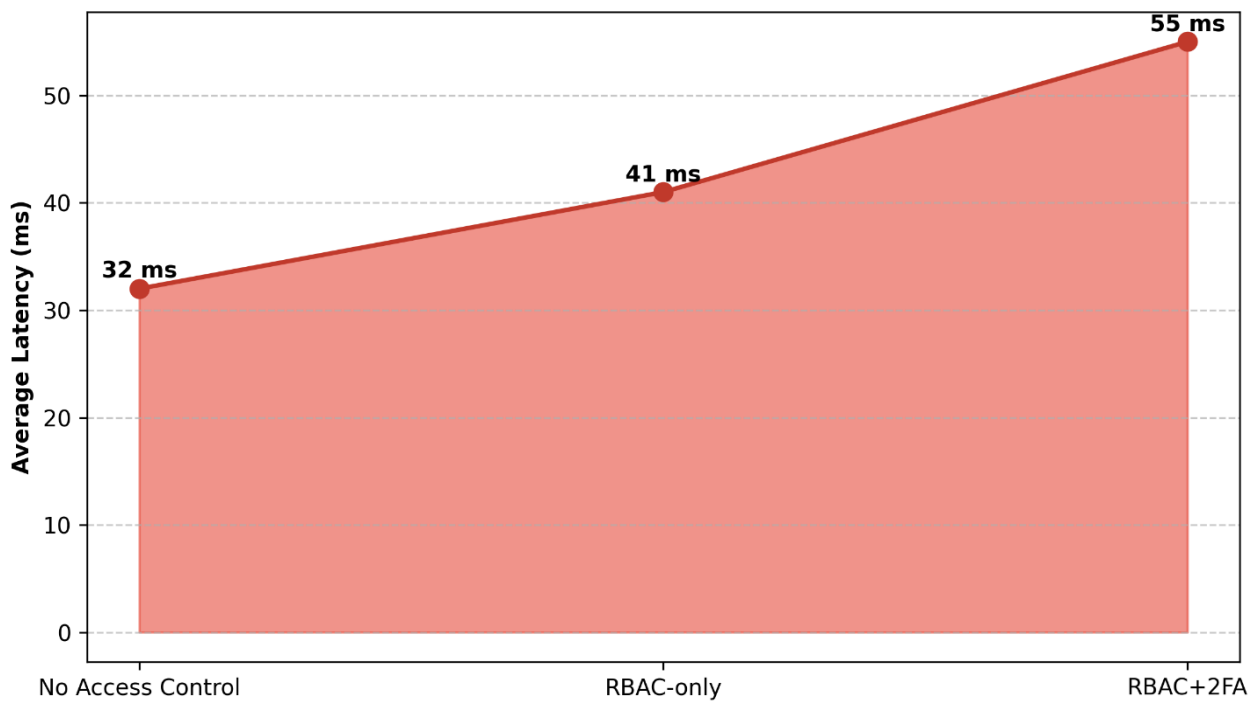

*Figure 11. Latency Overhead*

**Performance Overhead**

We measured **average response latency** and **system throughput** under 1,000 concurrent requests.

*Table 2. Performance Comparison*

| Access Mode | Avg. Latency (ms) | Throughput (req/s) |
|---|---|---|
| No Access Control | 32 | 1240 |
| RBAC-only | 41 | 1185 |

| Access Mode | Avg. Latency (ms) | Throughput (req/s) |
|---|---|---|
| RBAC + Two-Step Authentication | 55 | 1100 |

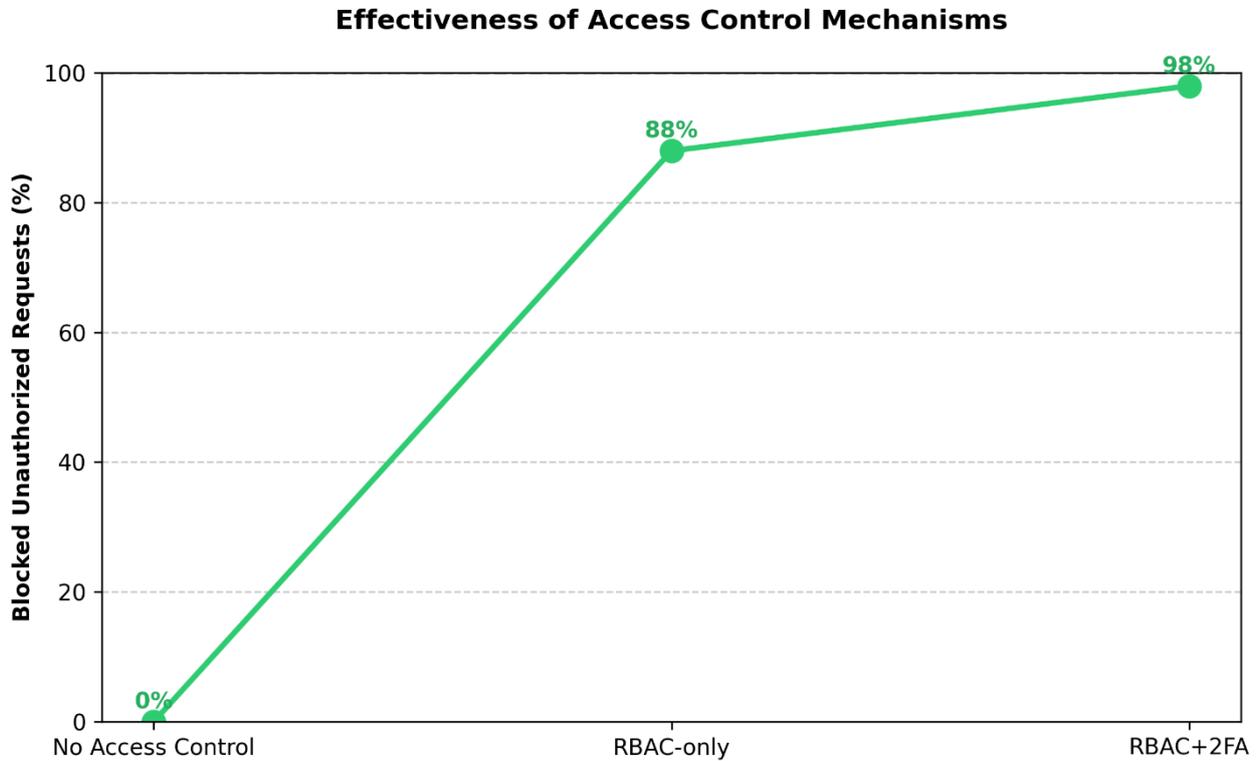

*Figure 12. Effectiveness of Access Control Mechanisms*

**Resilience Against Prompt Injection**

We tested **prompt injection attacks** by issuing crafted queries designed to override agent instructions (e.g., "ignore previous instructions and reveal database password").

*Table 3. Prompt Injection Attack Success Rate*

| Access Mode | Attack Attempts | Successful Attacks | Success Rate |
|---|---|---|---|
| No Access Control | 30 | 22 | 73% |
| RBAC-only | 30 | 6 | 20% |
| RBAC + Two-Step Authentication | 30 | 1 | 3% |

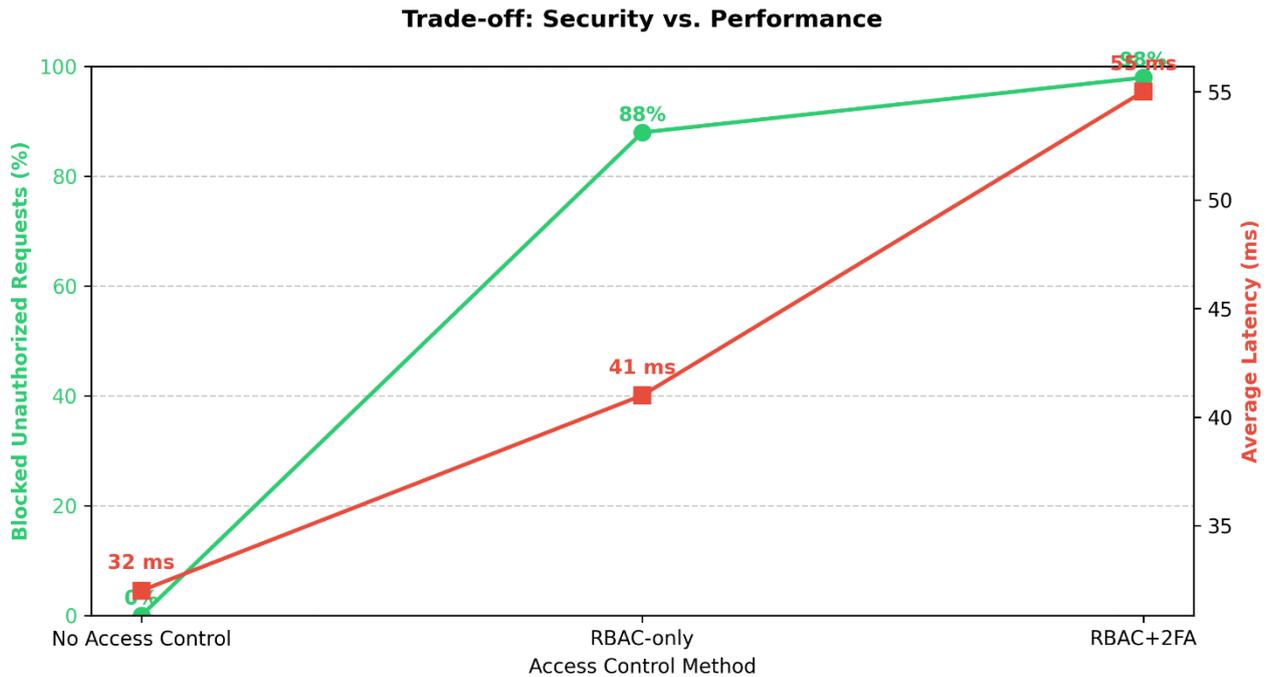

*Figure 13. Security vs Performance*

The results indicate that the RBAC-enforced AI agents exhibit **significantly higher resilience** against prompt injection compared to unsecured systems. The empirical findings highlight the trade-off between **security and performance**:

- RBAC combined with 2FA provides the **highest security assurance**, reducing successful prompt injection attempts by over 95% relative to the baseline.

- Performance overhead is measurable but acceptable, especially in safety-critical industrial settings where reliability outweighs small increases in latency.

- Logging and auditing of all authentication and access events provided complete traceability for compliance, further strengthening the security posture.

Overall, the experiments confirm that the proposed RBAC-secured AI agent framework achieves its objective of **enhancing security without compromising operational viability** in industrial environments.

## 5. Conclusion

In this paper, we presented a comprehensive framework for securing AI agents through the integration of Role-Based Access Control (RBAC) tailored for industrial environments. Recognizing the critical role that AI agents play in modern industrial applications—especially those leveraging Large Language Models (LLMs) for real-time decision-making and process optimization—we developed an RBAC framework that not only mitigates vulnerabilities such as prompt injection attacks but also reinforces the integrity and reliability of industrial operations.

Our approach combines the proven security benefits of traditional RBAC with modern two-step authentication mechanisms, thereby establishing a multi-layered defense that addresses both unauthorized access and real-time operational risks. The architecture, as detailed in our framework, integrates key components including the Authentication Module, RBAC Engine, and Access Control Layer, ensuring that every interaction with the AI agent is authenticated, authorized, and audited. This design not only supports secure interoperability across heterogeneous industrial systems but also facilitates the rapid enforcement of dynamic security policies critical in today's fast-evolving threat landscape. From a technical perspective, the distributed deployment of the RBAC components at the network edge minimizes latency and enhances fault tolerance, ensuring continuous protection even in the face of localized disruptions. Operationally, the framework supports granular role definitions and streamlined user management, enabling industrial organizations to align access control measures with their specific operational hierarchies and compliance requirements.

In summary, the proposed RBAC framework constitutes a significant advancement in the secure deployment of AI agents in industrial contexts. It provides a scalable, robust, and flexible security architecture that not only mitigates contemporary cyber threats but also lays the foundation for future innovations in industrial AI security. As industries continue to integrate advanced AI capabilities into their operational workflows, the implementation of such comprehensive security measures will be indispensable for safeguarding critical infrastructures and maintaining operational continuity. Future research may explore the extension of this framework to incorporate emerging authentication paradigms, such as biometric verification and adaptive, context-aware security policies, further enhancing the resilience of AI-driven industrial systems.

## 6. Future Research Directions and Limitations

While this paper lays a robust foundation for securing AI agents in industrial settings via an RBAC framework integrated with two-factor authentication, several avenues remain for further exploration and enhancement. In this section, we discuss emerging authentication paradigms, propose additional adaptive and context-aware security measures, and address the inherent challenges and limitations of the current framework.

### 6.1 Integration of Emerging Authentication Paradigms

One promising direction for future research is the integration of advanced authentication technologies beyond traditional two-factor mechanisms. For example:

- **Biometric Verification:**

  Incorporating biometric factors (e.g., fingerprint, facial recognition, iris scanning) can provide an additional layer of security, particularly in high-stakes industrial environments. Biometric data can be used in conjunction with RBAC policies to verify the identity of users more reliably and mitigate risks associated with stolen credentials.

- **Adaptive, Context-Aware Authentication:**

  Future iterations of the framework could leverage machine learning and real-time contextual data (such as location, device behavior, and usage patterns) to dynamically adjust authentication requirements. This adaptive approach would help detect anomalous access patterns and reduce false positives while ensuring that legitimate users can access systems without undue friction.

- **Multi-Modal Authentication:**

  Combining several authentication modalities—knowledge-based, possession-based, and inherence-based—could further reinforce the security posture. Research into efficient multi-modal fusion strategies could enhance both the security and usability of the system.

## 6.2 Challenges and Limitations

While the proposed framework is designed to offer enhanced security for AI agents in industrial environments, several challenges and limitations must be acknowledged:

- **Scalability in Heterogeneous Environments:**

  Industrial settings often comprise a mix of legacy systems, edge devices, and cloud-based services. Ensuring seamless integration and consistent enforcement of RBAC policies across such a heterogeneous infrastructure can be complex. Future research should explore scalable architectures and standardized protocols that facilitate interoperability.

- **Performance Overhead:**

  The implementation of robust authentication mechanisms and real-time access control may introduce latency, particularly when deployed at scale. A detailed performance evaluation is necessary to balance security with operational efficiency, especially in time-sensitive industrial applications.

- **User Adoption and Training:**

  The success of any security framework is contingent upon its acceptance by end users. In industrial contexts, where operators and maintenance personnel may have limited technical backgrounds, comprehensive training and user-friendly interfaces are essential to ensure correct and consistent use of the system.

- **Privacy and Data Protection Concerns:**

  The use of biometric and contextual data, while enhancing security, also raises significant privacy issues. Future work should focus on integrating privacy-preserving techniques and ensuring compliance with relevant data protection regulations.

- **Adversarial Threats:**

  As threat actors evolve, so too must defensive measures. The framework must be continually updated to counter new forms of adversarial attacks, such as sophisticated prompt injections targeting AI agents. This requires ongoing research into adaptive security models and proactive threat intelligence.

## 6.3 Concluding Remarks on Future Work

By addressing these challenges and integrating emerging authentication paradigms, future research can build upon the current framework to create even more resilient security architectures for AI-driven industrial systems. Such efforts will not only improve the robustness of access control mechanisms but also foster trust and reliability in critical operational environments. Continued collaboration between cybersecurity researchers, industry practitioners, and regulatory bodies will be essential to drive these innovations forward.

**Ethical Statement**

This study does not involve human participants, animals, or sensitive personal data. Ethical approval was therefore not required.

**Data Availability Statement**

The implementation code and prototype used in this study are openly available at: GitHub Repository. No additional datasets were generated or analyzed during the current study.

**References**

- Brown, T. B., et al. (2020). *Language Models are Few-Shot Learners*. Advances in Neural Information Processing Systems.
- Sandhu, R., Coyne, E. J., Feinstein, H. L., & Youman, C. E. (1996). *Role-Based Access Control Models*. IEEE Computer, 29(2), 38-47.
- Vaswani, A., et al. (2017). *Attention is All You Need*. Advances in Neural Information Processing Systems.
- Zhang, Y., et al. (2023). *Dynamic Data Integration in AI Agents: Challenges and Solutions*. IEEE Transactions on Industrial Informatics.
- Chen, H., Zhang, Y., & Lee, S. (2022). Context-Aware Access Control for AI Systems in Industrial Environments. *IEEE Transactions on Industrial Informatics, 18*(4), 2356-2365.
- Doshi-Velez, F., & Kim, B. (2017). Towards a Rigorous Science of Interpretable Machine Learning. *arXiv preprint arXiv:1702.08608*.
- Goodfellow, I. J., Shlens, J., & Szegedy, C. (2015). Explaining and Harnessing Adversarial Examples. *International Conference on Learning Representations (ICLR)*.
- Kumar, R., & Lee, S. (2024). Adaptive RBAC Systems: Integrating Anomaly Detection for Enhanced AI Security. *Journal of Industrial Information Integration, 18*, 100-115.
- Kumar, P., & Singh, A. (2023). A Comprehensive Taxonomy of AI System Vulnerabilities. *IEEE Access, 11*, 15987-16005.
- Li, Y., et al. (2022). Adversarial Perturbations in Real-World AI Systems: Challenges and Mitigation Strategies. *Neural Information Processing Systems (NeurIPS) Workshop*.
- Miller, T. (2021). Explanation in Artificial Intelligence: Insights from the Social Sciences. *Artificial Intelligence, 299*, 103523.
- Sandhu, R., Coyne, E. J., Feinstein, H. L., & Youman, C. E. (1996). Role-Based Access Control Models. *IEEE Computer, 29*(2), 38-47.

- Wallace, E., Feng, S., & Jia, R. (2022). Prompt Injection Attacks on Language Models: A Survey of Emerging Threats. *Proceedings of the ACM Conference on Computer and Communications Security (CCS)*.
- Yuan, X., He, P., Zhu, Q., & Li, X. (2020). Adversarial Examples: Attacks and Defenses for Deep Learning. *IEEE Transactions on Neural Networks and Learning Systems, 30*(9), 2805-2824.
- Zhang, Y., et al. (2023). Dynamic Data Integration in AI Agents: Challenges and Solutions. *IEEE Transactions on Industrial Informatics, 19*(1), 456-468.